\documentclass[lettersize,journal]{IEEEtran}
\usepackage{amsmath,amsfonts,amssymb}
\usepackage{enumerate}
\usepackage{algorithmic}
\usepackage{algorithm}
\usepackage{array}
\usepackage[caption=false,font=normalsize,labelfont=sf,textfont=sf]{subfig}
\usepackage{textcomp}
\usepackage{stfloats}
\usepackage{url}
\usepackage{verbatim}
\usepackage{graphicx}
\usepackage{cite}
\usepackage{color}
\usepackage{multirow}
\usepackage{booktabs}
\usepackage{lineno,hyperref}
\begin{document}
	
	\title{SAR-Net: Multi-scale Direction-aware SAR Network via Global Information Fusion}
	
	\author{Mingxiang Cao,
		Weiying Xie,~\IEEEmembership{Senior Member,~IEEE},	
		Jie Lei,~\IEEEmembership{Member,~IEEE},
		Jiaqing Zhang,
		Daixun Li,
		Yunsong Li,~\IEEEmembership{Member,~IEEE}
		\thanks{This work was supported in part by the National Natural Science Foundation of China under Grant 62071360. (Corresponding~authors: Weiying Xie; Jie Lei.)
			
			Mingxiang Cao, Weiying Xie, Jiaqing Zhang, Daixun Li and Yunsong Li are with the State Key
			Laboratory of Integrated Services Networks, Xidian University, Xi'an 710071,
			China (e-mail: mingxiangcao@stu.xidian.edu.cn; wyxie@xidian.edu.cn; jqzhang\_2@stu.xidian.edu.cn; ldx@stu.xidian.edu.cn;  ysli@mail.xidian.edu.cn). Jie Lei is at the School of Electrical and Data Engineering at the University of Technology Sydney(e-mail: jie.lei@uts.edu.au).}}
	
	\markboth{}%
	{Shell \MakeLowercase{\textit{et al.}}: A Sample Article Using IEEEtran.cls for IEEE Journals}
	
	
	\maketitle
	
	\begin{abstract}

		Deep learning has driven significant progress in object detection using Synthetic Aperture Radar (SAR) imagery. Existing methods, while achieving promising results, often struggle to effectively integrate local and global information, particularly direction-aware features. This paper proposes SAR-Net, a novel framework specifically designed for global fusion of direction-aware information in SAR object detection. SAR-Net leverages two key innovations: the Unity Compensation Mechanism (UCM) and the Direction-aware Attention Module (DAM). UCM facilitates the establishment of complementary relationships among features across different scales, enabling efficient global information fusion. Among them, Multi-scale Alignment Module (MAM) and distinct Multi-level Fusion Module (MFM) enhance feature integration by capturing both texture detail and semantic information. Then, Multi-feature Embedding Module (MEM) feeds back global features into the primary branches, further improving information transmission. Additionally, DAM, through bidirectional attention polymerization, captures direction-aware information, effectively eliminating background interference. Extensive experiments demonstrate the effectiveness of SAR-Net, achieving state-of-the-art results on aircraft (SAR-AIRcraft-1.0) and ship datasets (SSDD, HRSID), confirming its generalization capability and robustness.
	\end{abstract}
	
	
	\section{Introduction} \label{introduction}
	\IEEEPARstart{O}{BJECT} detection plays a pivotal role in remote sensing, with extensive applications in military operations, security surveillance, disaster assessment, and environmental monitoring\cite{ref1,ref2}. Synthetic Aperture Radar (SAR) object detection, a key technique in remote sensing, offers notable benefits in super-resolution imaging, hidden object detection, and all-weather surveillance, particularly in identifying aircraft and ships.
	
	In recent years, with the rapid development of deep learning, an increasing number of researchers have turned their attention to design object detection methods using Convolutional Neural Networks (CNNs), yielding notable results \cite{HRLEref13,HRLEref14,HRLEref15,HRLEref20,HRLEref21,oh2020spam,HRLEref22}. However, the unique challenges posed by SAR images, characterized by their multi-scale and directional nature, coupled with abundant background details, complicate object detection tasks. These factors particularly hinder the detection of small objects amidst background noise, making the design of effective multi-scale object detection networks for SAR images a complex task. Consequently, developing strategies to accurately determine the direction and position of aircraft or ships in SAR images, despite background interference, stands as a critical challenge in this field.
	
	Recent advancements in SAR image object detection have prominently featured deep learning approaches. Addressing the challenges of detecting objects in varying directions, Li \textit{et al}. \cite{ref8} developed the Spatial Frequency Feature Fusion Network. This innovative network utilizes a Feature Pyramid Network (FPN) \cite{FPN} to create spatial multi-scale features, specifically for ship objects, through a top-down hierarchical approach. Additionally, it employs a polar coordinate Fourier transform to attain rotation-invariant features in the frequency domain. In another notable work, Zhang \textit{et al}. \cite{ref7} introduced the Quad-Feature Pyramid Network (Quad-FPN), an architectural innovation consisting of four feature pyramid networks. This design effectively confronts issues such as background noise, multi-scale object representation, and the subtle features of small boats. Complementing these advancements, Li \textit{et al}. \cite{ref9} developed the Adjacent Feature Fusion (AFF) module, which strategically integrates local shallow features into adjacent layers. This selective and adaptive approach significantly enhances feature fusion, leading to improved representation of objects in SAR images.
	\begin{figure}[tpb]
		\centering
		\includegraphics[width=1.0\linewidth]{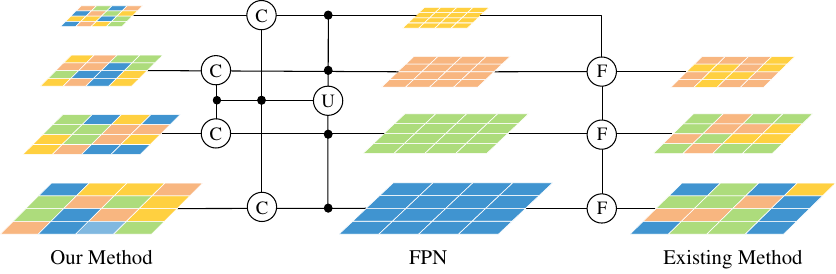}
		\caption{Comparison of information fusion and transmission between our method and existing methods. In the figure, U represents unity operation, C represents feature compensation, and F represents feature fusion. Our perspective enables each layer to focus on global information, while existing fusion methods only focus on local information, resulting in information loss.}
		\label{fig:1.1}
		\vspace{-0.1in}
	\end{figure}
	
Current research in SAR image object detection primarily focuses on utilizing FPN or its variants for feature fusion, offering a partial solution to multi-scale challenges \cite{ref7, ref9}. However, a significant limitation of these methods is their lack of global information integration at each layer. As illustrated in Fig. \ref{fig:1.1} (right side), local information flow between adjacent layers leads to a dilution of information across non-adjacent layers, resulting in information loss \cite{CVPR2019}. Recognizing the importance of global information, Bai \textit{et al}. \cite{ref10} proposed a Global Context-Guided Feature Balance Pyramid (GC-FBP), incorporating a global context module to bridge distant relationships between ship objects and their background. Yet, information flows similar to FPN are still used, and their approach of simply stacking context information onto the original feature map impedes effective interplay of information. Our method, depicted on the left side of Fig. \ref{fig:1.1}, diverges from traditional algorithms that primarily depend on straightforward concatenation of adjacent feature layers. Instead, we adopt a global information fusion strategy, allowing for feature integration across various scales. This approach not only enhances the effectiveness and efficiency of the fusion process but also minimally impacts information loss during transmission.
	
Traditional SAR object detectors, which predominantly use horizontal boxes, typically capture features of directional objects through an attention module \cite{guo2023break} or by transforming features into the frequency domain \cite{ref8}. However, these methods tend to neglect the correlation between the object and its contextual environment. This oversight often leads to ambiguous differentiation between the object and the background, with the resultant features being significantly noisy. Furthermore, these techniques generally overlook the crucial role of positional information. In contrast, our approach considers not only the object's features but also the surrounding context. By integrating directional and positional information and embedding channel attention, our approach offers a comprehensive solution.
	
	We have distilled the problem of SAR image object detection into a broader challenge: \textit{addressing the issues of multi-scale and directional objects in predominantly background-dominated images.} We base our considerations on three key premises: (1) the significant variation in object scales in SAR images. (2) the need to consider both shallow textural and deep semantic information of objects. (3) the directional nature of objects coupled with background interference. To address these challenges, we introduce a unity compensation mechanism, which includes three components: the Multi-scale Alignment Module (MAM), the Multi-level Fusion Module (MFM), and the Multi-feature Embedding Module (MEM). MAM initially unites multi-scale features across different spatial hierarchical levels. MFM then integrates overarching semantic cues before globally amalgamating them with intricate, fine-grained details, providing a comprehensive overview of the united features. Lastly, MEM compensates for any original feature to enrich their representational information. This multifaceted approach not only improves the detection of multi-scale objects but also significantly enhances information fusion capabilities.
	
	In addressing the third premise, we present a direction-aware attention module equipped with bidirectional attention polymerization capabilities. This innovative module employs deformable convolution and global pooling operations to extract
	object positional and directional information along the rows and columns of input features. It then generates a direction-aware attention map, which is effectively integrated with the original input features. This approach adeptly tackles the challenge of accurately discerning directional and positional information amidst intricate background interference. Our contributions can be summarized in three primary points:
	
	$\bullet$ We present SAR-Net, an innovative framework tailored specifically for object detection in SAR images. SAR-Net incorporates a unity compensation mechanism that globally unites multi-scale features of objects while compensating for the original feature. This ensures that global information is inherently integrated within features of different scales.
	
	$\bullet$ We design a direction-aware attention module with bidirectional attention polymerization capabilities. This module extracts object features while obtaining directional and positional information, enabling the identification of object rotation poses. Such a design is particularly effective in complex SAR images, where the directional attributes of objects are often obscured by significant background interference.
	
	$\bullet$ To validate the effectiveness, generalizability, and robustness of SAR-Net, we conduct comprehensive comparative and ablation experiments using three publicly available benchmark datasets for SAR aircraft and ship detection. The results of these experiments clearly show that our SAR-Net framework outperforms other state-of-the-art methods in all three datasets.
	
	The subsequent sections of this paper are structured as follows: Section II provides an overview of the evolution of object detection within the realm of SAR imagery, alongside a survey of relevant literature addressing multi-scale challenges. Section III details our proposed framework and modules. In Section IV, the experimental setup is described, and qualitative and quantitative results are reported. Finally, conclusions are drawn in Section V.
	
	\section{Related Work}
	\subsection{SAR Object Detection}
Object detection in SAR images has gained significant attention in recent years. Traditional detection algorithms, which follow the Constant False Alarm Rate (CFAR) principles \cite{CFAR}, operated based on a statistical model. In prior works \cite{2018adaptive}, \cite{2019ship}, researchers aimed to segregate objects from the background to emphasize disparities and streamline detection processes. Some other works \cite{2016modified, 2017improved, 2013improved, 2016inshore} were devoted to solving the problem of multi-scale object detection in high-resolution images.
Despite these achievements, traditional methods still face constraints in detecting small objects and heavily depend on manual design, posing challenges for robustness and limiting their applicability to different SAR datasets. Hence, there is a critical demand for more adaptable and automated approaches in SAR object detection.
	
Owing to the inherent capacity for autonomous feature learning in deep learning, researchers have seamlessly integrated object detection in SAR images with deep learning methodologies\cite{10384614, rs13214209, 9681317, 10412205, 9496207, 10522760}. Shifting focus to SAR ship detection and identification, Zhang \textit{et al}. \cite{2023joint} proposes a joint unsupervised super-resolution and SAR object detection network. This network augments object details while suppressing noise, providing valuable guidance to the detection network. In the realm of SAR aircraft detection and recognition, Zhao \textit{et al}. \cite{attentional} introduces an attention module, strategically fusing refined low-level texture features with high-level semantic features, which enhances aircraft detection rates by leveraging complementary information. Furthermore, Zhao \textit{et al}. \cite{pyramid} devises a multi-branch dilated convolution feature pyramid method. This innovative approach minimizes redundancy while accentuating critical aircraft features through the establishment of dense connections. In a related endeavor, Zhao \textit{et al}. \cite{2023multitask} put forth a multitask learning framework for object detection (MLDet) tailored for ship detection in SAR imagery. This end-to-end framework encompasses several tasks and had a good effect. The collaborative execution of these tasks contributes to an enhanced detection performance. Addressing the specific challenge of large-scale SAR images, Yang \textit{et al}. \cite{2023multi} holistically considers both pre-processing and post-processing in algorithm optimization, leading to an overall improvement in the effectiveness of object detection in large-scale SAR images. The comprehensive strategy employed by these studies underscores the effectiveness of deep learning paradigms in advancing the state-of-the-art in SAR object detection and recognition.
	
Although progress has been made, scale variation across object instances is still one of the most knotty problems in SAR object detection. 
	\begin{figure*}[htpb]
		\centering
		\includegraphics[width=1.0\linewidth]{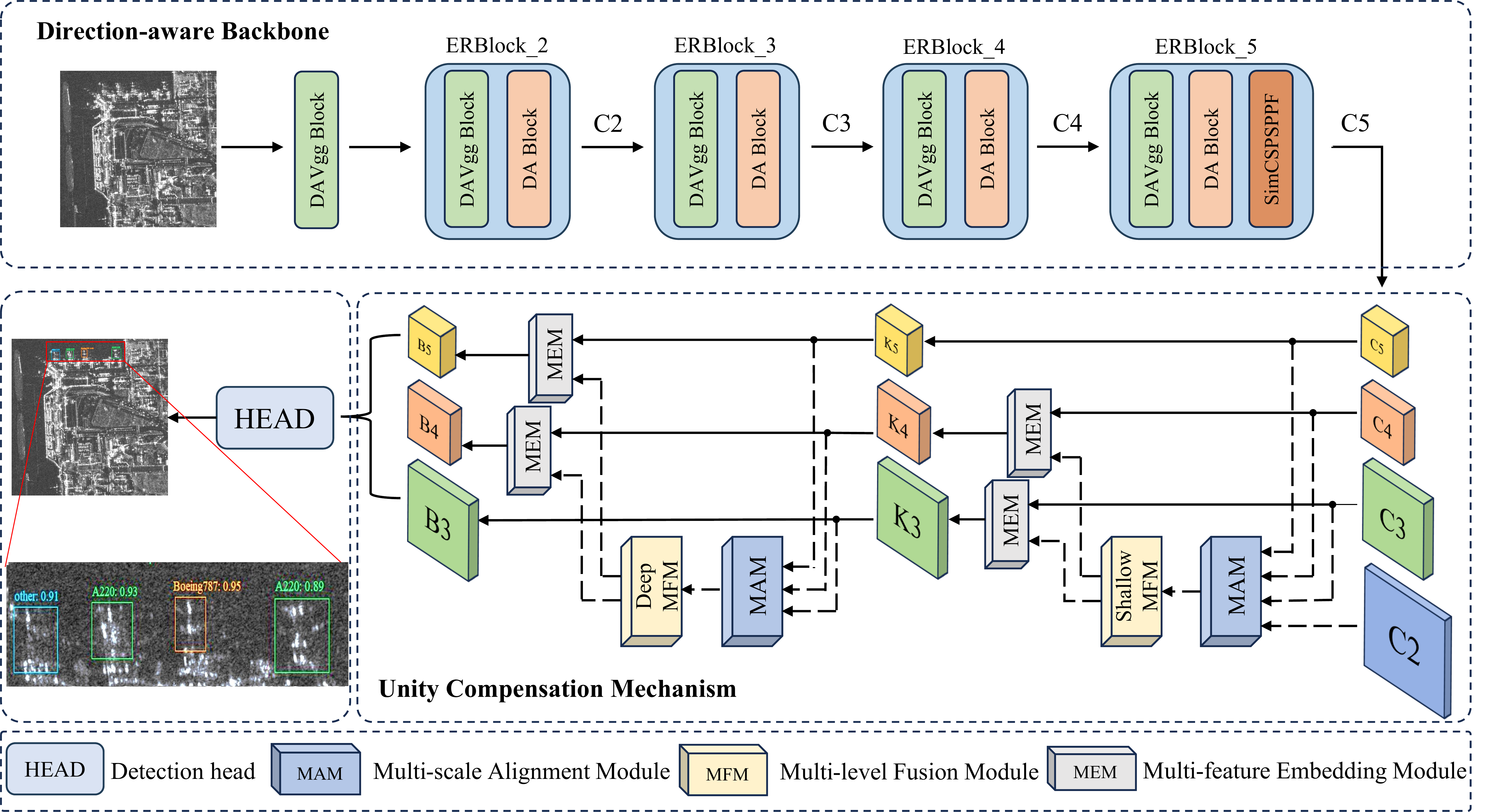}
		\caption{Overview of the proposed SAR-Net framework, here, the DAVgg Block embeds our designed direction-aware attention module, and the DA Block consists of DAVgg Blocks, combined to extract effective direction-aware information. $Ci$, $Ki$, and $Bi$ represent the multi-scale features at different stages.}
		\label{fig:3.1}
		\vspace{-0.1in}
	\end{figure*}
	\subsection{Multi-scale Object Detection}
In the realm of deep learning, the adoption of image pyramids initially emerged as a solution for tackling multi-scale challenges in object detection, driven by its intuitive appeal \cite{cheng2022tiny, zhang2019beyond, huang2020novel, li2022dense, zhou2023transformer}. Notable approaches like SNIP \cite{singh2018analysis} and SNIPER \cite{singh2018sniper} distinguished themselves by selecting specific scales for each resolution during multi-scale training. However, it's essential to note that these methods significantly increase inference time, creating a trade-off between accuracy and efficiency. Another approach is to build a feature pyramid inside the network, providing an approximate image pyramid and reducing computational costs, as seen in SSD \cite{liu2016ssd}. This enables the detection of multi-scale objects at each feature layer but can lead to decreased accuracy in detecting small objects. The issue arises from the low-level features themselves lacking semantic information, which is particularly crucial for small object detection.
	
	To address the semantic limitations in low-level features, the Feature Pyramid Network (FPN) \cite{FPN} is introduced. FPN offers an efficient structural design for merging multi-scale features via skip connections and local information transfer. Building upon FPN, the Path Aggregation Network (PANet) \cite{liu2018path} introduces a bottom-up path to enhance information fusion between different levels. Researchers at FSAF \cite{zhu2019feature} identify a flaw in FPN's heuristic-guided feature selection, highlighting the need for improvement. In contrast, FSAF focuses attention on regions of interest within each hierarchical feature level, aiming to extract features from a single level, but ignores valuable information between different hierarchical levels. The SA-FPN \cite{ni2019multi}, amalgamating Top-Down style FPN and Bottom-Up style FPN, combined the strengths of both, emerging as a more accurate module for perceiving scale variations. While previous methods integrate information at all levels through common operations like element-wise max or sum, these fusion mechanisms do not consider semantic differences among feature layers. EfficientDet \cite{tan2019efficientnet} addressed this limitation by introducing a repeatable module (BiFPN) to enhance information fusion efficiency between different levels. YOLO-ASFF \cite{liu2019learning} employed an adaptive spatial feature fusion technique, learning weights during training to filter out inconsistencies. Recognizing that the feature maps in a pyramid predominantly consist of single-level features, M2Det \cite{duan2019centernet} provided insights into this structure. Moreover, \cite{yang2023afpn} extended FPN with the Asymptotic Feature Pyramid Network (AFPN) to facilitate interaction across non-adjacent layers. To overcome FPN's limitations in detecting large objects, \cite{chen2021parallel} proposed a parallel FPN structure for object detection with bi-directional fusion, while \cite{jin2022you} introduced a refined FPN structure. SAFNet \cite{jin2020safnet} introduced Adaptive Feature Fusion and Self-Enhanced Modules. 
	
	However, some problems remain in the FPN structure. These methods all use sequential local fusion, which is an indirect interaction method and ignores the loss in the information flow process.  Inspired by \cite{CVPR2019}, all feature map layers present meaningful information for object detection. Therefore, designing a direct interaction and global fusion structure is of great significance for multi-scale object detection.  
	\section{Methods}
	This paper presents SAR-Net, an SAR object detection framework that incorporates a unity compensation mechanism and a direction-aware attention module. Built upon the YOLOv6 \cite{yolov6-3.0} architecture, SAR-Net consists of three components: backbone, neck, and head. Leveraging the advanced detection head of YOLOv6, we retain its head in the SAR-Net framework while focusing on innovations in the backbone and neck. Within this section, we first present the comprehensive architecture of SAR-Net, followed by an introduction to our proposed unity compensation mechanism, which addresses the issue of hindered information interaction due to local information fusion. Finally, we clarify the newly designed direction-aware attention module embedded in the backbone and neck, dedicated to acquiring object directional and positional information.
	\begin{figure}[htpb]
		\centering
		\includegraphics[width=1.0\linewidth]{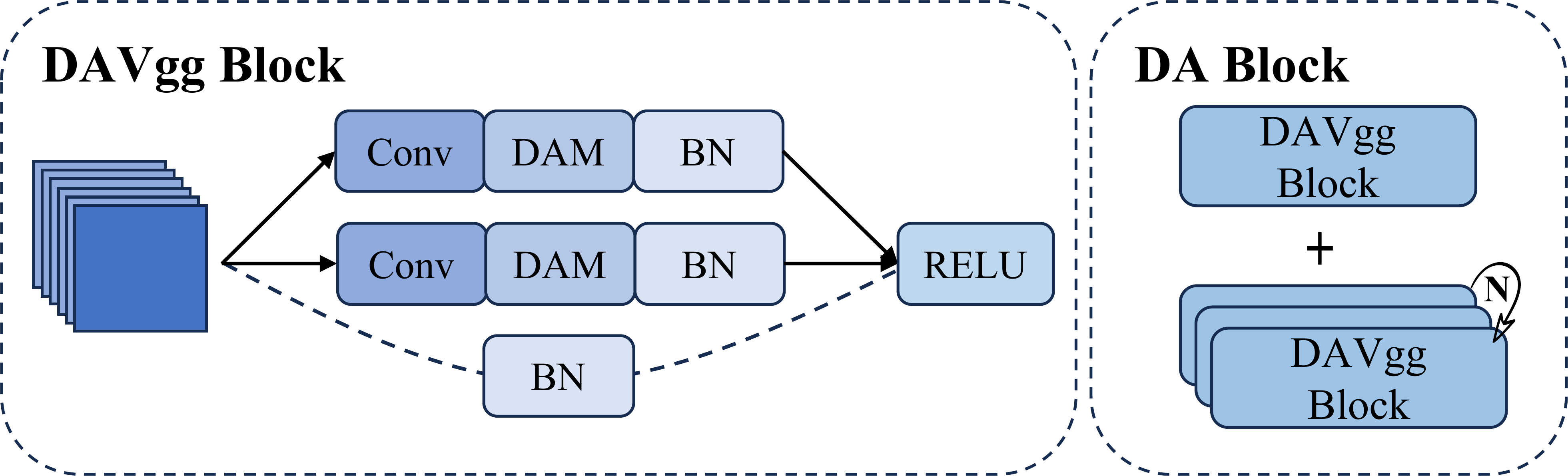}
		\caption{Overview of backbone components. DAM represents the Direction-aware Attention Module.}
		\label{fig:3.2}
		\vspace{-0.1in}
	\end{figure}
		\begin{figure}[htpb]
		\centering
		\includegraphics[width=1.0\linewidth]{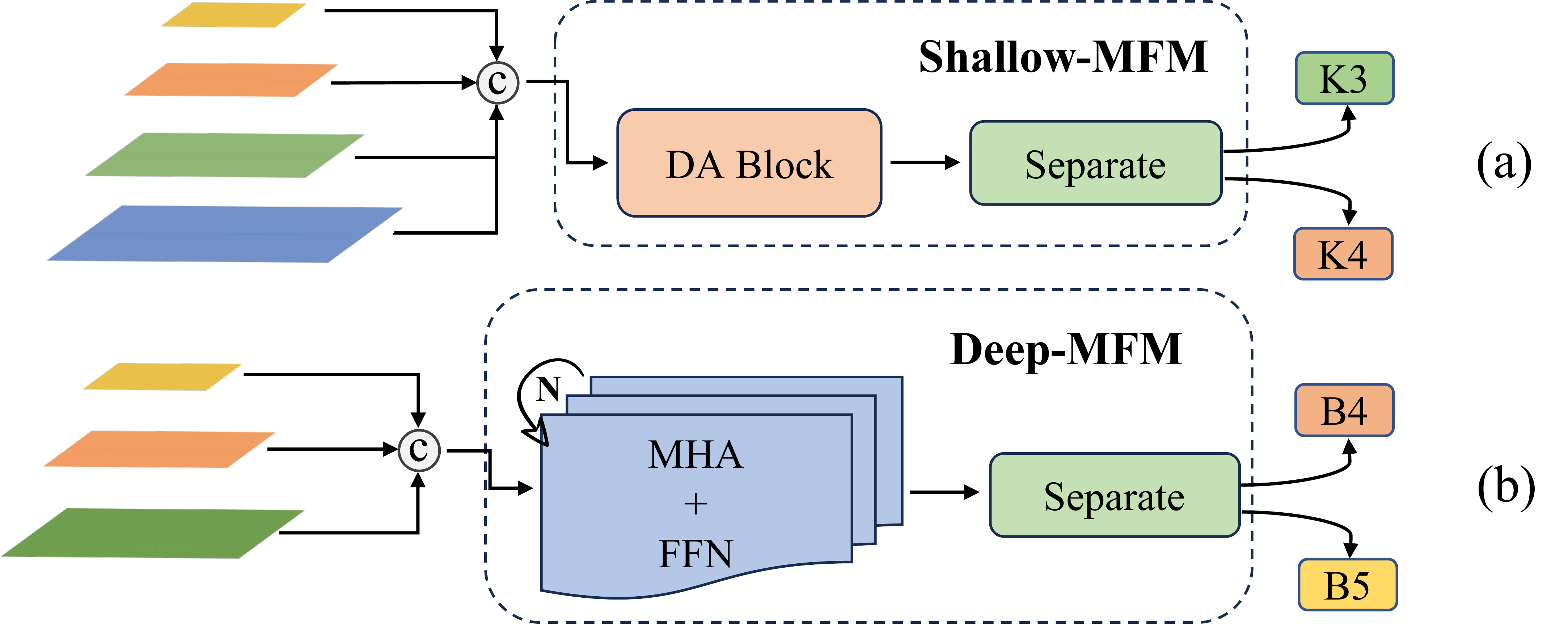}
		\caption{Multi-level Fusion Module at both shallow and deep levels, here, MHA and FFN denote Multi-Head Attention and Feed-Forward Network, respectively, with N of them combined in transformer blocks. The Multi-scale Alignment process is also depicted on the left side of the diagram, and the concatenation operation is performed along the channel dimension.}
		\label{fig:3.3}
		\vspace{-0.1in}
	\end{figure}
	\subsection{Overall Framework of SAR-Net}
	The comprehensive framework of SAR-Net is depicted in Fig. \ref{fig:3.1}, where the backbone and neck components are specifically tailored for SAR images. Specifically, considering that SAR image content is predominantly background, and detected objects are directional, using feature extraction networks designed for optical images directly cannot capture the directional and positional information of objects under complex background interference. To this end, we have designed a direction-aware feature extraction backbone network, as shown in Fig. \ref{fig:3.2}. It embeds our designed direction-aware attention module, reducing background interference while capturing fine-grained information, and providing effective assistance for the subsequent global fusion of multi-scale features in SAR images. Additionally, to better tackle the challenge of multi-scale in SAR images and avoid information dilution caused by traditional FPN local fusion, we propose a unity compensation mechanism. This mechanism comprehensively fuses low-level and high-level features across various scales, enabling each layer to focus on global information and providing robust prior knowledge for the detection head. 
	\subsection{Unity Compensation Mechanism}
	In our implementation, the unity and compensation process is facilitated by three modules: the Multi-scale Alignment module, the Multi-level Fusion module, and the Multi-feature Embedding module. These modules collaborate to unitedly compensate for features, we illustrate the detailed information in the following subsections. As depicted in Fig. \ref{fig:3.1}, the input to the neck includes the feature maps $C2$, $C3$, $C4$, and $C5$ extracted from the backbone, where $Ci\in\mathbb{R}^{N\times\tilde{C_{Ci}}\times H_{Ci}\times W_{Ci}}$. The batch size is represented as $N$, the number of channels as $C$, and the dimensions as $H\times W$. Additionally, the feature map sizes decrease progressively from $C2$ to $C5$, while the channel numbers double at each level.

	\subsubsection{Multi-scale Alignment Module}
	Given the crucial premise that both shallow texture and deep semantic information of objects in SAR images are important, we choose the outputs $C2$, $C3$, $C4$, and $C5$ from the backbone for global feature alignment. We consider two factors for aligning feature size: the computational complexity of global fusion and the necessity of preserving the fine-grained information extracted by the backbone. The former tends to favor smaller feature maps to reduce computational load, while the latter prefers larger feature maps to preserve detailed information. To balance these two aspects, the second smallest feature size is chosen as aligned size when four features need to be aligned while the smallest feature size is selected when aligning three features.
	
	We employ average pooling and bilinear interpolation operations to adjust the feature sizes to the aligned feature size, obtaining $\mathcal{F}_{global\_align}$. This module ensures effective aggregation of information through global alignment while preserving high-resolution features of small objects. The formula is as follows:
	\begin{equation}
		\label{equ:3.1}
		\mathcal{F}_{global\_align}=[AvgPool(Xi),Xj,Interpolate(Xk)]_{c},
	\end{equation}
	here, $\left[...\right]_{c}$ denotes the concatenation operation along the channel dimensions. $Xi$, $Xj$, and $Xk$ represent the different multi-scale features to be aligned at the same stage.
	\subsubsection{Multi-level Fusion Module}
	Due to the typically high resolution and rich detailed information in SAR images, small objects manifest as shallow texture details, while large objects exhibit deep semantic information. Performing fusion only once cannot fully preserve the information of objects at different scales, to augment the model's proficiency in detecting multi-scale objects, we devised two Multi-level Fusion modules: the Shallow Multi-level Fusion Module (Shallow-MFM) and the Deep Multi-level Fusion Module (Deep-MFM).
	
	\textbf{Shallow-MFM.} Given that convolutional operations can effectively extract local features and texture information, which is crucial for detecting small objects in SAR images, Shallow-MFM utilizes Direction-aware Attention Block (DA Block). As shown in Fig. \ref{fig:3.3} (a), these blocks ensure the comprehensive capture and fusion of global detailed information in SAR images, followed by a separation operation. The formula is as follows:
	\begin{equation}
		\label{equ:3.2}
		\mathcal{F}_{global\_fuse}=DA \; Block(\mathcal{F}_{global\_align}),
	\end{equation}
	\begin{equation}
		\label{equ:3.3}
		\mathcal{F}_{emb\_K3},\mathcal{F}_{emb\_K4}=Seperate(\mathcal{F}_{global\_fuse}).
	\end{equation}
	
	As shown in Fig. \ref{fig:3.1}, for better control of the information propagation and fusion process, we only separate two features and compensate them onto the input features $C3$ and $C4$. This is because $C5$ contains the position and feature information of large objects, while the compensated features encompass more information about small objects. Simultaneously, the feature information from $C2$ has already been sufficiently utilized, and we no longer use it in the subsequent fusion process.
	
	\textbf{Deep-MFM.} Due to the advantages of transformers in handling global relationships and long-distance dependencies, and considering that large objects require more global information for accurate detection and localization, we employ a transformer fusion module in Deep-MFM. As depicted in Fig. \ref{fig:3.3} (b), it comprises a stack of $N$ transformer blocks, with each block containing Multi-Head Attention (MHA) and a Feed-Forward Network (FFN). We follow the configuration for Multi-Head Attention from \cite{levit}, and this part of the process is expressed by the formula:
	\begin{equation}
		\label{equ:3.4}
		\mathcal{F}_{global\_fuse}=Transformer(\mathcal{F}_{global\_align}),
	\end{equation}
	\begin{equation}
		\label{equ:3.5}
		\mathcal{F}_{emb\_B3},\mathcal{F}_{emb\_B4}=Seperate(\mathcal{F}_{global\_fuse}).
	\end{equation}
	
	Simultaneously, in constructing the FFN, we utilize depth-wise convolution, which can be considered as an implicit channel attention mechanism, enhancing the local connections of the transformer block. It is represented as follows:
	\begin{equation}
		\label{equ:3.6}
		\mathcal{F}_{FFN}=CB_{2}(\psi(DW_{conv}(CB_{1}(X)))),
	\end{equation}
	here, $CB_{i}$ denotes the $i$-th convolution and batch normalization operation, $DW_{conv}$ represents Depth-Wise convolution, and $\psi$ is the ReLU activation function.
	\begin{figure}[htpb]
		\centering
		\includegraphics[width=1.0\linewidth]{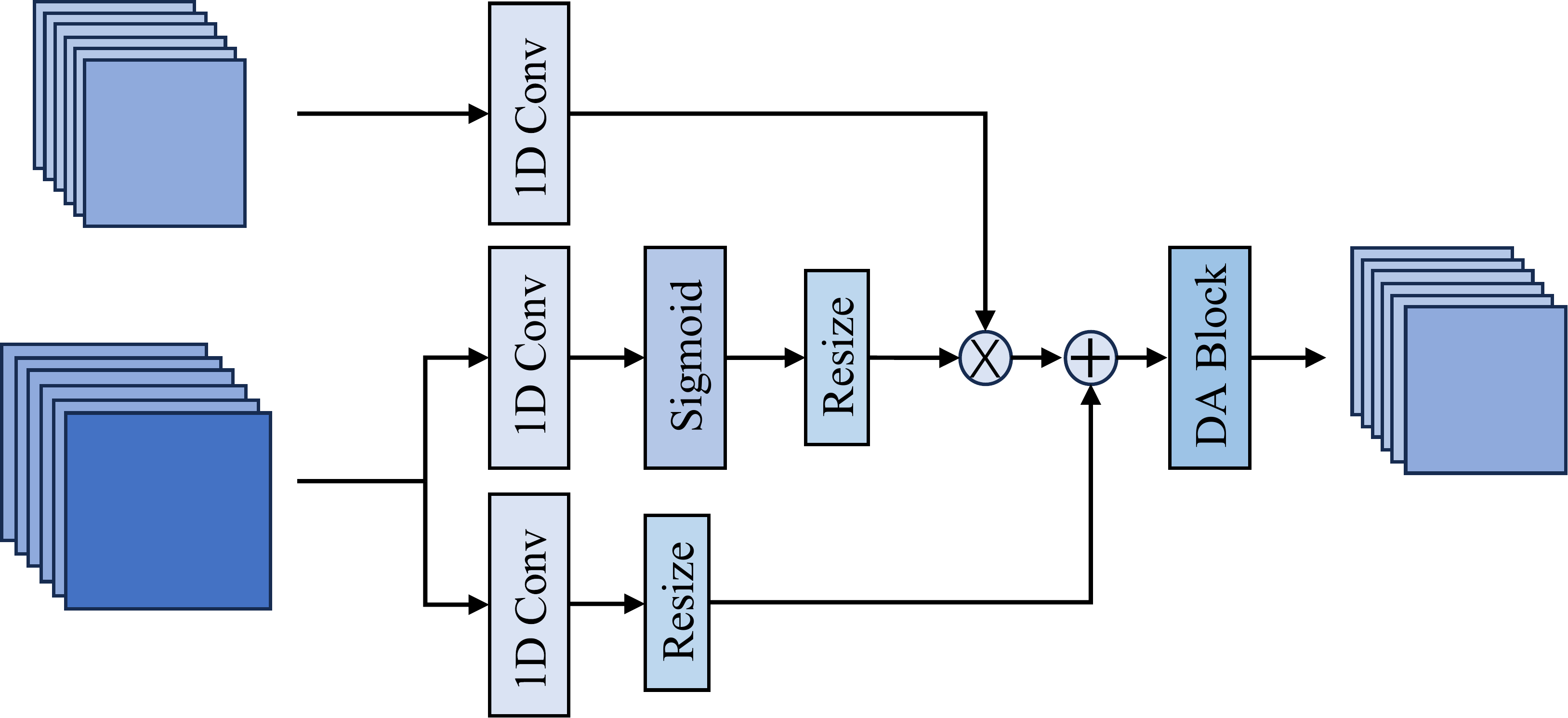}
		\caption{Architecture of Multi-feature Embedding Module, where Resize represents the average pooling operation or bilinear interpolation.}
		\label{fig:3.4}
		\vspace{-0.1in}
	\end{figure}
	
	Before merging the output features, to retain information about small objects in $K3$, we compensate the separated features onto the intermediate features $K4$ and $K5$, providing the detection head with output features rich in information about both small and large objects.

	\subsubsection{Multi-feature Embedding Module}
	Inspired by the fusion architecture with two branches in \cite{seaformer}, to embed global information more effectively at different levels, we employ attention operations for information fusion, as illustrated in Fig. \ref{fig:3.4}. In particular, we feed local information (pertaining to features at the operating level) and globally embedded information (generated by MFM) into two separate 1D convolutions for feature transformation. The Sigmoid function is utilized to generate attention weights from global information, which are then applied to weight the local information. Additionally, for further compensation, a residual connection is employed. This is aimed at achieving deep fusion and enhancement of global and local information. Finally, the DA Block is used for further feature extraction and fusion.
	\subsection{Direction-aware Attention Module}
	Research on object detection in SAR images has demonstrated that attention mechanisms can notably enhance model performance. However, existing attention mechanisms often concentrate solely on positional information, frequently neglecting directional cues. This results in a lack of a method that seamlessly integrates both directional and positional information and embeds them into the attention map. In response to this, we introduce the direction-aware attention module.
	
\begin{figure}[htpb]
		\centering
		\includegraphics[width=1.0\linewidth]{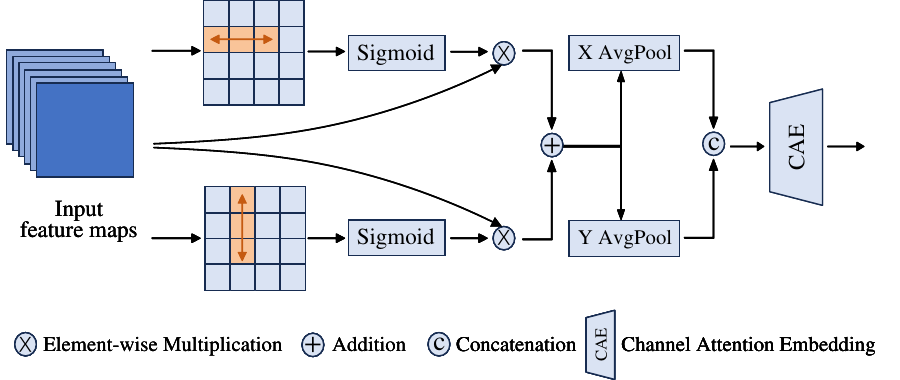}
		\caption{Architecture of Direction-aware Attention Module. The concatenation operation is performed along the spatial dimension.}
		\label{fig:3.5}
		\vspace{-0.1in}
	\end{figure}

	As depicted in Fig. \ref{fig:3.5}, it can be regarded as a means of enhancing the feature map's representational capacity, obtaining direction-aware feature maps through two steps: Direction-aware Generation and Channel Attention Embedding. These steps contribute to facilitating the subsequent extraction of object features. Below, we provide a detailed description of these two steps.
	
	\subsubsection{Direction-aware Generation}
Due to the dominance of background in SAR image content and the presence of interference areas such as docks, airports, and terminals, directly locating and detecting objects becomes challenging. Additionally, SAR images often exhibit the edge contours of objects, reflecting distinct directional features, while extracting detailed information within the objects is challenging. Therefore, leveraging the stability and distinctiveness of object direction information can effectively reduce the difficulty of localization \cite{su2022sii}. 
	
	Based on the analysis above, deformable convolution, capable of adapting to and describing object deformations well, is employed. We perform deformable convolution along one direction (either the row or column direction) on the input features, generating two directional attention maps. These attention maps are then used to modulate the input features, and it is expressed as:
	\begin{equation}
		\label{equ:3.7}
		X_{direction}=(\varphi(DConv_{row}(X)+DConv_{column}(X)))\otimes X,
	\end{equation}
here, $X$ represents the input features, $\varphi$ represents the Sigmoid function, $\otimes$ represents element-wise matrix multiplication, as the dimensions of the matrices before and after the operation are equal. This bidirectional convolution attention mechanism facilitates the extraction of directional information from low-level features in two directions, thereby guiding the refinement of input features.

	To further leverage the directional attention maps and avoid capturing only local contextual information, we forego using large-kernel convolutions and instead employ two parallel $1D$ feature encoding approaches. These methods independently aggregate input features in the vertical and horizontal directions, resulting in two separate localization-aware feature maps, as illustrated below:
	\begin{equation}
		\label{equ:3.8}
		Z^h=\frac{1}{W}\sum_{i=0}^{W}X_{direction}(h,i)\quad h=0,1,...,H,
	\end{equation}
	\begin{equation}
		\label{equ:3.9}
		Z^w=\frac{1}{H}\sum_{j=0}^{H}X_{direction}(j,w)\quad w=0,1,...,W,
	\end{equation}
	the output feature maps $Z^h$ and $Z^w$ encapsulate global statistical information along the height and width directions, effectively aggregating multi-directional global contextual features of the input. This fusion of spatial positional information proves beneficial for subsequent detection tasks.
	\subsubsection{Channel Attention Embedding}
	The aforementioned process enables a global receptive field and encodes precise directional awareness. To fully leverage this information and capture inter-channel relationships, we first align the two feature maps along the spatial dimensions. Subsequently, they are fed into the convolutional transformation function CBR, yielding:
	\begin{equation}
		\label{equ:3.10}
		g=CBR([Z^{h},Z^{w}]_{s}),
	\end{equation}
	where $\left[...\right]_{s}$ denotes the concatenation operation along the spatial dimensions, and CBR represents the sequence of convolution, batch normalization, and RELU activation function. It encodes the features with both horizontal and vertical directional information after reintegration. Subsequently, we decompose the concatenated features into $g^h$ and $g^w$ along the spatial dimension. Two $1D$ convolutions are employed to restore the feature dimensions, and finally, the weight information from the Sigmoid function is applied to the input features, resulting in:
	\begin{equation}
		\label{equ:3.11}
		Y=X*\varphi\left(Conv_h(g^h)\right)*\varphi(Conv_w(g^w)),
	\end{equation}
where * represents tensor product.  	
	
	In SAR object detection, we leverage the directional nature of SAR image objects by applying attention mechanisms to input features from two directions, extracting features in both horizontal and vertical directions. Simultaneously, we use average pooling and convolution to reduce the spatial scale and channel dimensions of intermediate features, reducing computational complexity in the embedding process. Compared to traditional methods, this encoding process enables our directional attention to more accurately identify objects of interest and achieve more efficient object detection.
	\subsection{Loss Function}
	The overall loss of our network consists of two components \cite{yolov6-3.0}: the loss of object classification $L_{cls}$ and the loss related to bounding box regression $L_{reg}$. It can be expressed as:
	\begin{equation}
		\label{equ:3.12}
		L_{total}=L_{cls}+L_{reg},
	\end{equation}
	we adopt the VariFocal Loss (VFL) \cite{varifocalnet} as the classification loss $L_{cls}$. VFL balances the learning signals between positive and negative samples by considering their varying levels of importance. The bounding box regression loss $L_{reg}$ comprises IOU loss and probability loss, as shown below:
	\begin{equation}
		\label{equ:3.13}
		L_{reg}=\lambda_{iou}L_{iou}+\lambda_{dfl}L_{dfl},
	\end{equation}
	here, $\lambda_{iou}$ and $\lambda_{dfl}$ are hyperparameters. We choose the advanced SIOU \cite{siou} as the IOU loss $L_{iou}$. Due to the consideration of ambiguity and uncertainty in the data without introducing additional strong priors, the Distribution Focal Loss (DFL) \cite{DFL} contributes to improving box localization accuracy. Additionally, we add a necessary loss $L_{dfl}$, which accounts for the bounding box regression loss.
	\section{Experiments and Analysis}
	To verify the superiority of SAR-Net, we perform extensive experiments on three public benchmark SAR object detection datasets, a fine-grained aircraft detection dataset SAR-AIRcraft-1.0 \cite{zhirui2023sar} and two ship detection datasets SSDD \cite{zhang2021sar}, HRSID \cite{wei2020hrsid}. Furthermore, we benchmarked our proposed SAR-Net against SOTA detection methods, and the results demonstrated the superior performance of SAR-Net.
	\begin{table}[htpb]
		\vspace{-0.1in}
		\setlength{\tabcolsep}{7mm}
		\renewcommand{\arraystretch}{1.5}
		\centering
		\caption{The Total of Instance Objects of Different Categories on the SAR-AIRcraft-1.0}
		\begin{tabular}{cccc}
			\specialrule{1.3pt}{0pt}{0pt}
			Categories      & Train      & Test     & Total  \\ \specialrule{1.3pt}{0pt}{0pt}
			A330    & 278    & 31  & 309    \\
			A320/321     & 1719   & 52 & 1771 \\
			A220 & 3270   & 460  & 3730    \\       
			ARJ21    & 825   & 362   &  1187   \\
			Boeing737     & 2007 &  550  & 2557  \\
			Boeing787 & 2191  & 454 & 2645  \\
			other  & 3223  & 1041 & 4264  \\ \specialrule{1.3pt}{0pt}{0pt}
			Total & 13513  &2950 & 16463   \\
			\specialrule{1.3pt}{0pt}{0pt}
		\end{tabular}
		\label{tab:4.1}
		\vspace{-0.1in}
	\end{table}
	\begin{figure*}[htpb]
		\vspace{-0.3in}
		\centering
		\includegraphics[width=1.0\linewidth]{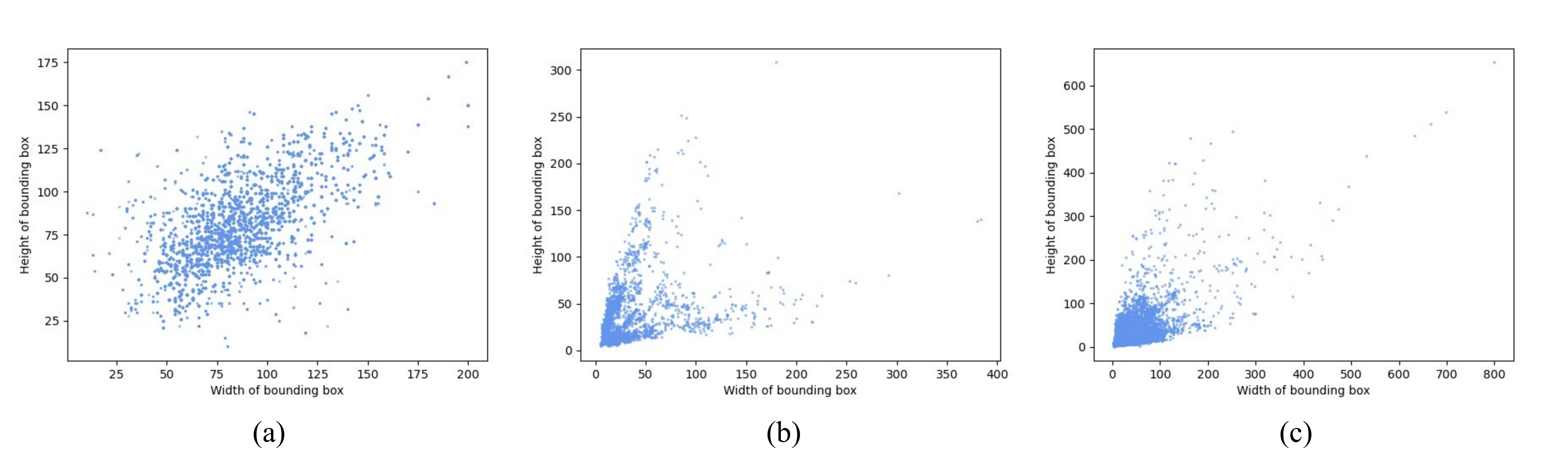}
		\caption{Bounding box sizes distribution of three SAR Datasets, where (a), (b), and (c) represent SAR-AIRcraft-1.0, SSDD, and HRSID datasets respectively.}
		\label{fig:4.1}
		\vspace{-0.1in}
	\end{figure*}
	\subsection{Dataset and Implementation Details}
	To meticulously assess the performance and generalizability of SAR-Net framework, we executed a lot of verification experiments on a recently released fine-grained multi-class SAR aircraft object detection dataset and two commonly used SAR ship object detection datasets.
	\begin{table}[htpb]  
		\vspace{-0.1in}
		\setlength{\tabcolsep}{0.2mm}
		\renewcommand{\arraystretch}{1.5}
		\centering
		\caption{Ablation Study of Three Modules on Unity Compensation Mechanism with MAM Using Concatenation Operation}
		\begin{tabular}{cccc|cccc}
			\specialrule{1.3pt}{0pt}{0pt}
			MAM&S-MFM & D-MFM & MEM & mAP$_{50}$(\%) & F1-score(\%) & Params(M) & GFLOPs  \\ \specialrule{1.3pt}{0pt}{0pt}
			&&  &  &86.5 &81.3 &4.6 &17.8  \\
			$\checkmark$  &  & &&85.9 &81.9 &4.0  &15.5 \\
		$\checkmark$  &	$\checkmark$  &  &  &87.1 &82.9 &4.2 &16.5  \\
		$\checkmark$  &	&  $\checkmark$ &  &87.0 &82.5 &4.7  &15.8 \\
		$\checkmark$  &	&  & $\checkmark$ &86.8 &82.7 &4.8  &17.5   \\  
		$\checkmark$  &	$\checkmark$&&$\checkmark$  &87.5 &83.3 &5.0 &18.5    \\
		$\checkmark$  &	&$\checkmark$&$\checkmark$  &87.3 &82.9 &5.4 &17.8    \\    
		$\checkmark$  &	$\checkmark$&$\checkmark$ &  &87.8 &83.5 &4.9 &16.8    \\
		$\checkmark$  &	$\checkmark$ & $\checkmark$ & $\checkmark$ &\bf{88.5} &\bf{84.6} &5.6 &18.8 \\
			\specialrule{1.3pt}{0pt}{0pt}
		\end{tabular}
		\label{tab:4.2}
		\vspace{-0.1in}
	\end{table}

	\subsubsection{SAR-AIRcraft-1.0}
	Images acquired from the Gaofen-3 satellite are screened to form the SAR-AIRcraft-1.0 dataset. The polarization method is single polarization, with a spatial resolution of $1m$, and utilizing the spotlight imaging mode. Considering the size of the airport and the number of parked aircraft, the dataset mainly uses image data of three civil airports: Shanghai Hongqiao Airport, Beijing Capital Airport, and Taiwan Taoyuan Airport, including a total of $800\times800$, $1000\times1000$, $1200\times1200$ and $1500\times1500$. 4 different sizes, a total of 4,368 pictures and 16,463 aircraft object instances. The specific categories of aircraft include A220, A320/321, A330, ARJ21, Boeing737, Boeing787, other. The total of instance objects of different categories is shown in Table \ref{tab:4.1}, where other represents aircraft instances that do not belong to the other 6 categories. The size distribution of the aircraft object slices in the dataset spans a wide range. As shown in Fig. \ref{fig:4.1} (a), some objects are smaller than $50\times50$, and some aircraft objects are larger than $100\times100$. The overall object size is multi-scale. In addition, in this paper, we use the default dataset partitioning by \cite{zhirui2023sar} and resize all images to $800\times800$ pixels. To ensure fairness, we adopt the above settings when compared with other methods. 
			\begin{table}[htpb]  
		\vspace{-0.1in}
		\setlength{\tabcolsep}{0.2mm}
		\renewcommand{\arraystretch}{1.5}
		\centering
		\caption{Ablation Study of Three Modules on Unity Compensation Mechanism with MAM Using Addition Operation}
		\begin{tabular}{cccc|cccc}
			\specialrule{1.3pt}{0pt}{0pt}
			MAM&S-MFM & D-MFM & MEM & mAP$_{50}$(\%) & F1-score(\%) & Params(M) & GFLOPs  \\ \specialrule{1.3pt}{0pt}{0pt}
			&&  &  &86.5 &81.3 &4.6 &17.8  \\
			$\checkmark$  &  & &&84.2 &80.7 &4.1  &15.8 \\
		$\checkmark$  &	$\checkmark$  &  &  &86.8 &82.3 &4.5 &16.8  \\
		$\checkmark$  &	&  $\checkmark$ &  &86.8 &82.2 &5.0  &16.1 \\
		$\checkmark$  &	&  & $\checkmark$ &86.6 &82.3 &5.1  &17.7   \\  
		$\checkmark$  &	$\checkmark$&&$\checkmark$  &87.2 &83.0 &5.3 &18.7    \\
		$\checkmark$  &	&$\checkmark$&$\checkmark$  &86.9 &82.6 &5.6 &18.1    \\     
		$\checkmark$  &	$\checkmark$&$\checkmark$ &  &87.1 &82.5 &5.2 &17.0    \\
		$\checkmark$  &	$\checkmark$ & $\checkmark$ & $\checkmark$ &\bf{88.0} &\bf{84.4} &5.9 &19.1 \\
			\specialrule{1.3pt}{0pt}{0pt}
		\end{tabular}
		\label{tab:4.10}
		\vspace{-0.1in}
	\end{table}
	\begin{figure}[htpb]
		\centering
		\includegraphics[width=1.0\linewidth]{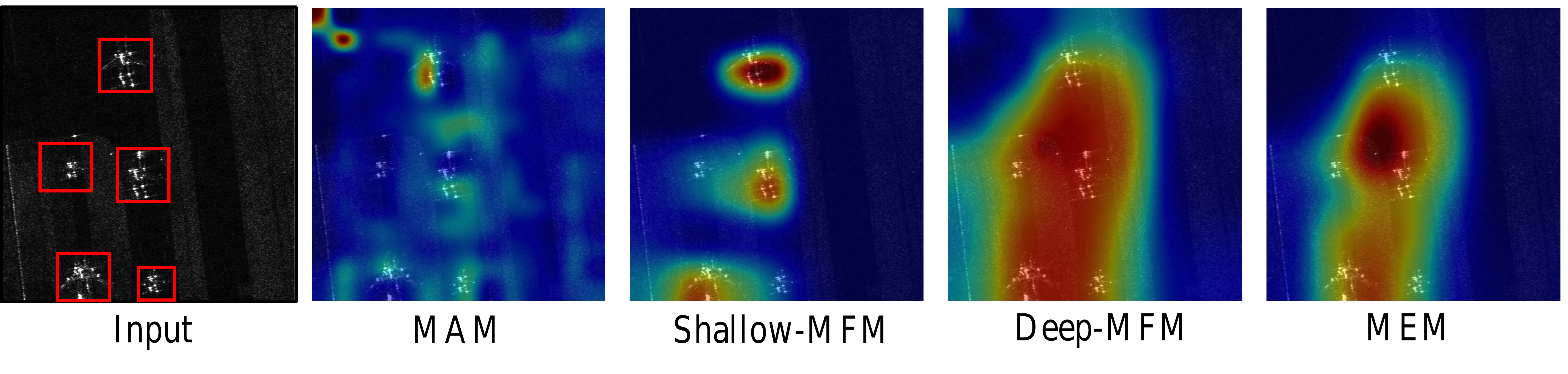}
		\caption{Layercam \cite{jiang2021layercam} visualization of different modules on Unity Compensation Mechanism.}
		\label{fig:4.5}
		\vspace{-0.1in}
	\end{figure}

	\subsubsection{SSDD}
	The SSDD dataset comprises data primarily sourced from three sensors. Its image resolution exhibits a wide range of variations. Comprising 1,160 SAR images and 2,456 ship objects, on average, each image contains 2.12 ship objects. The sizes of each ship object are shown in Fig. \ref{fig:4.1} (b). For standardized benchmarking, the images with numbers ending in 1 and 9 in the dataset form the testset, ensuring an 8:2 ratio division with the training set. To facilitate fair comparisons with other methods, all images are resized to $640\times640$ pixels in all experiments, maintaining consistent data distribution.
		\begin{table}[htpb]  
		\vspace{-0.1in}
		\setlength{\tabcolsep}{1.5mm}
		\renewcommand{\arraystretch}{1.5}
		\centering
		\caption{Research on the Effectiveness and Optimal Embedding Locations of Direction-aware Attention Module}
		\begin{tabular}{c|cccc}
			\specialrule{1.3pt}{0pt}{0pt}
			Settings & mAP$_{50}$(\%) & F1-score(\%) & Params(M) & GFLOPs  \\ 
			\specialrule{1.3pt}{0pt}{0pt}
			Baseline  &86.5 &81.3 &4.6 &17.8    \\
			+Row Attention &87.2 &82.5  &4.9 &18.8    \\
			+Column Attention &86.9 &82.4  &4.9 &18.8    \\
			DAM w/o CAE  &87.4 &83.3  &4.9 &19.1  \\
			DAM w/o DG  &86.8 &82.4  &4.7 &18.4  \\
			Behind RELU&87.4 &83.1  &4.8 &18.6    \\     
			Behind ERBlock&87.1 &81.9  &4.8 &18.3    \\
			DAM&\bf{87.8} &\bf{83.4}  &5.0 &19.7    \\
			\specialrule{1.3pt}{0pt}{0pt}
		\end{tabular}
		\label{tab:4.3}
		\vspace{-0.1in}
	\end{table}

	\subsubsection{HRSID}
	To further assess the ship detection capabilities of SAR-Net, we conducted experiments on HRSID, which is a large-scale, high-resolution SAR images dataset designed for ship object detection tasks. The HRSID dataset encompasses SAR images with diverse properties, including variations in sea area, polarization, sea condition, resolution, and coastal port scenarios. As depicted in Fig. \ref{fig:4.1} (c), the dataset encompasses objects at a wide range of scales, annotated with bounding boxes within various environments. Panoramic images with resolutions ranging from $1m$ to $5m$ are cropped to a uniform size of $800\times800$. In total, the HRSID dataset consists of 5,604 cropped SAR images, containing 16,951 ships. These images are distributed in an 6.5:3.5 ratio division with the training set and test set. This dataset diversity ensures comprehensive evaluation and validation of our proposed method's performance under various conditions and scenarios. 
		\begin{table}[htpb]  
		\vspace{-0.1in}
		\setlength{\tabcolsep}{1.2mm}
		\renewcommand{\arraystretch}{1.5}
		\centering
		\caption{Collaborative Research Within the Framework, UC Represents Unity Compensation}
		\begin{tabular}{cc|cccc}
			\specialrule{1.3pt}{0pt}{0pt}
			UC Mechanism &DAM & mAP$_{50}$(\%) & F1-score(\%) & Params(M) & GFLOPs  \\ 
			\specialrule{1.3pt}{0pt}{0pt}
			& &86.5 &81.3 &4.6 &17.8    \\
			\checkmark& &88.5 &84.6 &5.6 &18.8    \\
			&\checkmark &87.8 &83.4  &5.0 &19.7    \\
			\checkmark& \checkmark&\bf{90.2} &\bf{85.6}  &6.0 &20.8    \\    
			\specialrule{1.3pt}{0pt}{0pt}
		\end{tabular}
		\label{tab:4.4}
		\vspace{-0.1in}
	\end{table}
	
	Our proposed framework has been conducted using PyTorch and is executed on a workstation equipped with an NVIDIA A100 GPU. During the training phase, we set the batch size to 32, and the learning rate is 0.02, which gradually reduces to 0.0002 following a cosine decay schedule, and train for a total of 300 epochs to ensure it fully converged. Standard Stochastic Gradient Descent (SGD) is used as the optimizer, and the parameters are all at default values. For input images during training, we only used mosaics augment, discarding Hue Saturation Values (HSV) and random flipping as they are not beneficial for SAR images. 
	\begin{table}[htpb]  
		\vspace{-0.1in}
		\setlength{\tabcolsep}{2.0mm}
		\renewcommand{\arraystretch}{1.5}
		\centering
		\caption{Comparison with Different FPN Variants}
		\begin{tabular}{c|cccc}
			\specialrule{1.3pt}{0pt}{0pt}
			NECK & mAP$_{50}$(\%) & F1-score(\%) & Params(M) & GFLOPs  \\ 
			\specialrule{1.3pt}{0pt}{0pt}
			FPN\cite{FPN}&87.4 &83.3 &4.5 &17.6    \\   
			PAFPN\cite{wang2019panet}&87.2 &82.8 &4.8 &18.1    \\
			AFPN\cite{yang2023afpn}&85.2 &80.7 &4.6 &18.1    \\   	   
			RepBi-PAN\cite{yolov6-3.0}&86.5 &81.3 &4.6 &17.8   \\    
			UC Mechanism&\bf{88.5} &\bf{84.6} &5.6 &18.8    \\
			\specialrule{1.3pt}{0pt}{0pt}
		\end{tabular}
		\label{tab:4.5}
		\vspace{-0.1in}
	\end{table}
	\subsection{Accuracy Metrics}
	The accuracy assessment involves evaluating the concordance and distinctions between the detection outcome and the reference label. Key accuracy metrics such as recall, precision, and mAP (mean Average Precision) are employed to assess the performance of the compared methods. The computations for precision and recall metrics are delineated as follows:
	\begin{equation}
		\label{equ:4.1}
		Precision=\frac{TP}{TP+FP},
	\end{equation}
	\begin{equation}
		\label{equ:4.2}
		Recall=\frac{TP}{TP+FN},
	\end{equation}
	in the equations, the terms true positive (TP) and true negative (TN) represent accurate predictions, while false positive (FP) and false negative (FN) denote incorrect outcomes. Precision and recall are associated with commission and omission errors, respectively. The F1-score assesses both precision and recall rates, and it can be expressed as:
	\begin{equation}
		\label{equ:4.3}
		F1\textendash score=2\times\frac{Precision \cdot Recall}{Precision+Recall}.
	\end{equation}
	
	The mAP serves as a holistic indicator derived by averaging the values of Average Precision (AP) and utilizes an integral approach to compute the area under the Precision-Recall curve across all categories. The mAP calculation can be expressed as:
	\begin{equation}
		\label{equ:4.4}
		mAP=\frac{AP}{N}=\frac{\int_{0}^{1}p(r)dr}{N},
	\end{equation}
	in the formula, $p$ represents Precision, $r$ represents Recall, and $N$ stands for the number of categories. In addition, model parameter sizes and GFLOPs are also important indicators used to evaluate the model, which are related to the implementation and practical application of the model.
	\begin{table*}[htpb]
		\vspace{-0.1in}
		\setlength{\tabcolsep}{1.8mm}
		\renewcommand{\arraystretch}{1.8}
		\centering
		\caption{Comparison with Other Methods on SAR-AIRcraft-1.0. The Accuracy of Each category is Represented by mAP with an Intersection Over the Union Threshold of 0.5}
		\begin{tabular}{c|ccccccccccc}
			\specialrule{1.3pt}{0pt}{0pt}
			Method&A220&A320/321&A330&ARJ21&Boeing737&Boeing787&other& mAP$_{50}$(\%) & F1-score(\%) & Params(M) & GFLOPs  \\ 
			\specialrule{1.3pt}{0pt}{0pt}
			Faster R-CNN\cite{ren2015faster}&77.3&97.0&86.4&74.0&59.5&74.0&71.2&77.2&74.6&41.0&142.7\\ 
			Cascade R-CNN\cite{cai2018cascade}&75.2&97.7&88.6&79.5&55.0&69.6&69.8&76.9&72.8&69.1&185.8\\
			YOLOv6n\cite{yolov6-3.0}&\bf{92.6}&97.5&96.2&81.4&78.3&80.6&78.9&86.5&81.3&4.6&17.8\\
			YOLOv8n\cite{Jocher_YOLO_by_Ultralytics_2023}&90.2&97.8&96.5&84.3&81.7&79.9&83.5&87.7&83.0&3.1&13.5   \\
			SKG-Net\cite{fu2021scattering}&66.0&79.9&79.2&66.1&64.8&70.4&72.1&71.2&65.9  &31.3&103.6\\
			SA-Net\cite{zhirui2023sar}&80.3&94.3&88.6&78.6&59.7&70.8&71.3&77.7&-&-&-   \\
			MLSDNet\cite{chang2023mlsdnet}&85.1&96.9&91.5&83.2&71.7&72.1&78.4&82.7&77.9&1.4&6.7   \\
			HRLE-SARDet\cite{zhou2023hrle}&90.6&98.1&96.7&82.6&78.9&79.1&82.5&86.9&84.0&3.0&14.2   \\
			FBR-Net\cite{fu2020anchor}&88.1&95.7&94.4&82.2&79.6&77.8&81.4&85.6&80.1&32.4&141.3   \\
			AFSar\cite{wan2021afsar}&89.2&96.8&95.5&83.3&80.7&78.9&82.5&86.7&81.7&3.6&14.9   \\
			RT-DETR\cite{zhao2023detrs}&90.1&97.7&96.3&84.3&81.5&79.9&83.4&87.6&82.6&31.9&190.2   \\
			CCDN-DETR\cite{zhang2024ccdn}&89.7&97.3&96.0&83.8&81.2&79.4&83.0&87.2&82.4&23.6&163.7   \\
			SAR-Net&91.0&\bf{98.5}&\bf{96.9}&\bf{92.6}&\bf{86.6}&\bf{82.4}&\bf{83.6}&\bf{90.2}&\bf{85.6}&6.0&20.8   \\  
			\specialrule{1.3pt}{0pt}{0pt}
		\end{tabular}
		\label{tab:4.6}
		\vspace{-0.1in}
	\end{table*}
	\begin{table}[htpb]  
		\vspace{-0.1in}
		\setlength{\tabcolsep}{1.3mm}
		\renewcommand{\arraystretch}{1.5}
		\centering
		\caption{Comparison with the Latest Object Detection Methods and SAR Ship Detection Methods on SSDD}
		\begin{tabular}{c|cccc}
			\specialrule{1.3pt}{0pt}{0pt}
			Method & mAP$_{50}$(\%) & F1-score(\%) & Params(M) & GFLOPs  \\ 
			\specialrule{1.3pt}{0pt}{0pt}
			Faster R-CNN\cite{ren2015faster}&96.5 &95.3 &41.0 &91.3    \\  
			Cascade R-CNN\cite{cai2018cascade}&97.0 &95.5 &69.1 &118.9    \\   	   
			YOLOv6n\cite{yolov6-3.0}&97.9 &94.9 &4.6 &11.4    \\
			YOLOv8n\cite{Jocher_YOLO_by_Ultralytics_2023}&98.1 &95.6 &3.1 &8.7    \\
			LPEDet\cite{feng2022lightweight}&97.5 &93.8 &5.6 &18.3    \\
			MLSDNet\cite{chang2023mlsdnet}&97.2 &93.7 &1.4 &4.3    \\
			HRLE-SARDet\cite{zhou2023hrle}&97.6 &94.8 &3.0 &9.1    \\
			YOLOv8m\cite{Jocher_YOLO_by_Ultralytics_2023}&98.4 &96.8 &25.4 &78.9    \\
			FBR-Net \cite{fu2020anchor}&94.1 &92.8 &32.4 &94.0    \\
			AFSar \cite{wan2021afsar}&97.1 &94.7 &3.6 &9.9    \\
			RT-DETR \cite{zhao2023detrs}&97.5 &95.7 &31.9 &125.3    \\
			CCDN-DETR \cite{zhang2024ccdn}&97.2 &95.6 &23.6 &109.1   \\
			SAR-Net&\bf{98.8} &\bf{97.2} &6.0 &13.7 \\
			\specialrule{1.3pt}{0pt}{0pt}
		\end{tabular}
		\label{tab:4.7}
		\vspace{-0.1in}
	\end{table}
	\subsection{Ablation Study}
	To comprehensively and detailedly analyze the effectiveness of SAR-Net, we executed extensive ablation experiments with different framework designs and internal modules, and all experiments were performed on the SAR-AIRcraft-1.0 dataset. The best-performing results are highlighted in bold.
		\begin{table}[htpb]  
		\vspace{-0.1in}
		\setlength{\tabcolsep}{1.3mm}
		\renewcommand{\arraystretch}{1.5}
		\centering
		\caption{Comparison with the Latest Object Detection Methods and SAR Ship Detection Methods on HRSID}
		\begin{tabular}{c|cccc}
			\specialrule{1.3pt}{0pt}{0pt}
			Method & mAP$_{50}$(\%) & F1-score(\%) & Params(M) & GFLOPs  \\ 
			\specialrule{1.3pt}{0pt}{0pt}
			Faster R-CNN\cite{ren2015faster}&81.0 &74.2 &41.0 &142.7    \\  
			Cascade R-CNN\cite{cai2018cascade}&82.7 &81.3 &69.1 &185.8    \\   	   
			YOLOv6n\cite{yolov6-3.0}&91.2&87.3&4.6 &17.8    \\
			YOLOv8n\cite{Jocher_YOLO_by_Ultralytics_2023}&91.4&87.7&3.1 &13.5    \\
			LPEDet\cite{feng2022lightweight}&89.9&86.2 &5.6 &28.6    \\
			MLSDNet\cite{chang2023mlsdnet}&89.1&84.5&1.4&6.7    \\
			HRLE-SARDet\cite{zhou2023hrle}&90.8 &86.6 &3.0 &14.2    \\
			CMFT\cite{he2023cross}&89.6 &85.9 &-&-    \\
			YOLOv8m\cite{Jocher_YOLO_by_Ultralytics_2023}&92.8&89.1 &25.4 &123.1    \\
			FBR-Net \cite{fu2020anchor}&89.6 &85.1 &32.4 &141.3    \\
			AFSar \cite{wan2021afsar}&91.3 &87.2 &3.6 &14.9    \\
			RT-DETR \cite{zhao2023detrs}&91.9 &88.1 &31.9 &190.2  \\
			CCDN-DETR \cite{zhang2024ccdn}&91.6 &87.9 &23.6 &163.7     \\
			SAR-Net&\bf{93.4} &\bf{89.8} &6.0 &20.8 \\
			\specialrule{1.3pt}{0pt}{0pt}
		\end{tabular}
		\label{tab:4.8}
		\vspace{-0.1in}
	\end{table}
	
	\subsubsection{Impact of Unity Compensation Mechanism}
	To study the effectiveness of each module in the unity compensation mechanism and verify the correctness of uniting deep and shallow layers' global information, we executed a lot of ablation experiments on four of the modules and measured the performance with mAP$_{50}$ and F1-score, a common indicator in object detection. In addition, Parameter sizes and GFLOPs are also shown in Table \ref{tab:4.2} and Table \ref{tab:4.10}. We use YOLOv6n \cite{yolov6-3.0} as the baseline for ablation experiments to embed the modules of our unity compensation mechanism one by one. The feature alignment process is included in all experiments except the baseline. In Table \ref{tab:4.2}, we conducted MAM using concatenation operation, where S-MFM represents Shallow-MFM and D represents Deep-MFM. It can be seen that only when the MAM module is used, the accuracy is lower than the baseline. This is because it integrates semantic information and detailed texture information from different layers, but it lacks further fusion of object-specific information. In addition, other experiments have achieved performance improvements without a substantial increase in computational overhead, which proves the efficacy of our global information fusion idea. In particular, when the MEM module is not used, we use simple addition to compensate for the original features. 
	
To explore a new alignment pattern, we conducted MAM using addition operation. It is worth noting that both concatenation and addition operations require aligning features from each layer to have the same dimensions. As can be seen from Table \ref{tab:4.10}, the overall trends of the other modules are similar to those in Table \ref{tab:4.2}. However, due to the addition operation combining rich texture and semantic information directly from different layers, it made the information redundant and resulted in difficulties in subsequent fusion compensation operations, leading to information loss and performance degradation. Additionally, features from different layers have varying channel numbers, requiring channel alignment during the addition operation, which increases parameter and computation complexity.
	
To visually demonstrate the progressive effect of modules within the Unity Compensation Mechanism, we utilized Layercam for visualizing feature maps. As depicted in Fig. \ref{fig:4.5}, the MAM aligns features from different layers but does not perform fusion operations, resulting in a chaotic feature map with scattered focus points across the entire image. However, it still retains features from different hierarchical levels for subsequent fusion compensation operations. After passing through Shallow-MFM, detailed texture information is fused and compensated back to the original features, enabling the model to focus on each object with texture information. Subsequently, with Deep-MFM, semantic information is fused and compensated onto the texture information, allowing the model to focus on object areas without losing texture information. Finally, the MEM embeds all features, further narrowing down the focus areas, which aids in model detection. The intuitive visualization results validate that the Unity Compensation Mechanism enhances feature integration while maximizing the retention of semantic and texture information.

	\subsubsection{Impact of Direction-aware Attention Module}
	To prove the effectiveness of the proposed direction-aware attention module, we executed a series of detailed experiments, as shown in Table \ref{tab:4.3}. We split the attention modules in two directions and embedded them in the YOLOv6n backbone separately, both received performance improvements. To explore the roles of the two modules within DAM, we conducted experiments by using only one module at a time on the baseline. Here, w/o CAE represents without Channel Attention Embedding, and w/o DG represents without Direction-aware Generation. Both experiments showed improvement compared to the baseline. However, due to the absence of crucial bidirectional feature vector extraction, DAM w/o DG showed minimal performance enhancement, merely reallocating attention among channels. Equally noteworthy, after generating the direction-aware attention map, omitting the embedding of the attention map among channels also resulted in performance degradation. Therefore, the collaboration of these two modules enables the model to concentrate more effectively on the directional areas of objects. 
	
	In addition, we tried to embed DAM in different positions of the backbone. Except for the embedding position shown in Fig. \ref{fig:3.2}, we also tried to move the DAM behind the RELU and each ERBlock, and obtained 0.9\% and 0.6\% performance improvement respectively with a small increase in parameters. Exhaustive experiments allowed us to determine the most suitable embedding position. These experimental results show that the acquisition of direction-aware information is significantly helpful for SAR images containing directional objects.
		\begin{figure*}[htpb]
		\centering
		\includegraphics[width=1.0\linewidth]{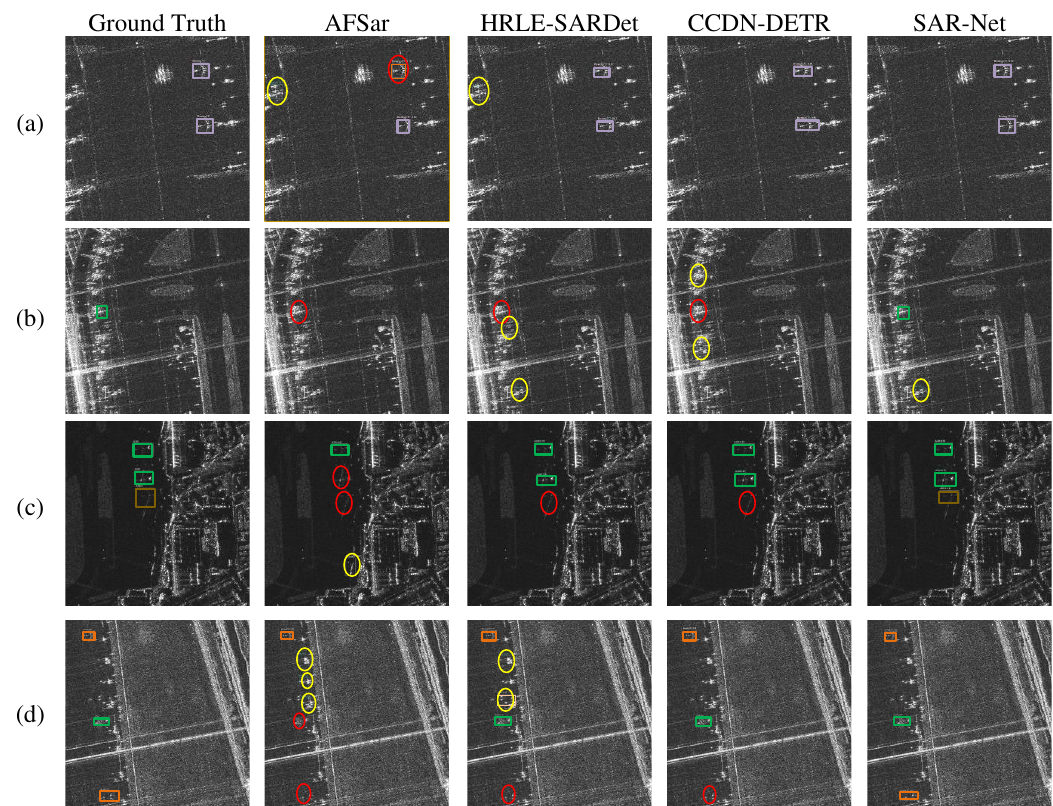}
		\caption{Visualization results of object detection under different methods on SAR-AIRcraft-1.0, (a)-(d) depict four randomly selected images, where differently colored rectangles represent aircraft of various categories. To enhance clarity, we have boldened the object boxes. Yellow ellipses indicate false positives, while red ellipses represent instances of false negatives and misclassifications.}
		\label{fig:4.2}
		\vspace{-0.1in}
	\end{figure*}
	
	\subsubsection{Collaboration of the Framework}
	The effectiveness of the components that make up SAR-Net has been demonstrated, and to further explore the synergy between them, we combine them and find the desired results. For the unity compensation mechanism, the backbone that can extract direction-aware information gives full play to the advantages of global fusion and enriches the information of multi-scale features. For the DAM-embedded backbone, the unity compensation mechanism retains detailed information and enhances the efficiency of information fusion and transmission. As shown in Table \ref{tab:4.4}, after combining the two, their collaborative complementarity is fully utilized and performance is further improved by 0.4\%.

	\subsubsection{Comparison of Different FPN Architecture}
	We choose the classic FPN structure and its latest variants AFPN \cite{yang2023afpn} and RepBi-PAN \cite{yolov6-3.0} for comparison. We conduct the experiments based on YOLOv6n and the results are illustrated in Table \ref{tab:4.5}. A significant accuracy disparity exists despite the parameters being similar. As we discuss in the section \ref{introduction}, sequential local fusion will lead to information loss, especially for SAR images with large object scale variants. Top-down and bottom-up structures will lose a lot of detailed information and semantic information before detection. Global information fusion and compensation preserve the object's multi-scale information to the maximum extent, thereby enabling the unity compensation mechanism to deliver optimal performance.
	\subsection{Comparisons with Previous Methods}
To showcase the effectiveness and robustness of our proposed method, we compared it with both classic and recent object detection methods on three datasets (SAR-AIRcraft-1.0, SSDD, HRSID). Additionally, we included comparisons with methods specifically designed for SAR images, such as those based on CNN and Transformer architectures. With the exception of the unreleased and non-detailed network structure of the comparative method, we successfully reproduce all other comparative methods. All methods are compared and evaluated from four aspects: mean Average Precision, F1-score, parameter size, and GFLOPs. The experimental parameter settings for each dataset remain consistent.

	\subsubsection{Comparison Results on SAR-AIRcraft-1.0}
	We use the SAR-AIRcraft-1.0 dataset to train and test several common benchmark detection methods, including two-stage detectors: Faster R-CNN \cite{ren2015faster}, Cascade R-CNN \cite{cai2018cascade}, and one-stage detectors: YOLOv6 \cite{yolov6-3.0}, YOLOv8 \cite{Jocher_YOLO_by_Ultralytics_2023}. In addition to this, we also selected three latest SAR aircraft detection methods: SKG-Net \cite{fu2021scattering}, SA-Net \cite{zhirui2023sar} and MLSDNet \cite{chang2023mlsdnet}, and some formidable model, HRLE-SARDet\cite{zhou2023hrle}, RT-DETR \cite{zhao2023detrs}, CCDN-DETR \cite{zhang2024ccdn}, FBR-Net \cite{fu2020anchor}, AFSar \cite{wan2021afsar}. Most of the methods are also applied to SSDD and HRSID. Table \ref{tab:4.6} shows the detection results under different methods.
	
			\begin{table}[htpb]
		\vspace{-0.1in}
		\setlength{\tabcolsep}{1.0mm}
		\renewcommand{\arraystretch}{1.5}
		\centering
		\caption{Comparison with the Latest Oriented SAR Ship Object Detection Methods on SSDD and HRSID}
		\begin{tabular}{c|cccc}
			\specialrule{1.3pt}{0pt}{0pt}
			\multirow{2}{*}{Method}  &\multicolumn{2}{c}{SSDD} &\multicolumn{2}{c}{HRSID} \\
			& mAP$_{50}$(\%) & F1-score(\%) & mAP$_{50}$(\%) & F1-score(\%)  \\ 
			\specialrule{1.3pt}{0pt}{0pt}
			Yue\cite{yue2023precise}&90.7 &88.3 &86.5 &83.9    \\  
			Li\cite{li2022oriented}&96.5 &94.8 &86.3 &83.2    \\   	   
			Zhou\cite{zhou2022pvt}&93.5 &91.6 &87.8 &84.7    \\
			KeyShip\cite{ge2023keyship}&90.7 &89.1 &83.7 &82.2    \\
			LPST-Det\cite{yang2024lpst}&93.9 &91.2 &87.8 &85.4    \\
			SAR-Net&\bf{98.8} &\bf{97.2} &\bf{93.4} &\bf{89.8} \\
			\specialrule{1.3pt}{0pt}{0pt}
		\end{tabular}
		\label{tab:4.9}
		\vspace{-0in}
	\end{table}
	
	In the Faster R-CNN method, the accuracy is 77.2\%, which shows to a certain extent that the SAR-AIRcraft-1.0 dataset has a certain detection difficulty because aircraft objects of different categories have similar structures and sizes, and the same category also presents multi-scale characteristics. As illustrated in Table \ref{tab:4.6}, SAR-Net achieves the highest performance in nearly every category, primarily due to its integration of global information, and on this basis, we only add a small amount of computational overhead, but the performance has been greatly improved. Considering the advanced nature of YOLOv8, we also experimented with its variants from small to large versions. We observed that the performance fluctuated within a range of 0.3\%, accompanied by a notable increase in computational cost. In addition, the detection accuracy of each category has certain differences. For example, A320/321 performs best in different algorithms compared to other categories. This is mainly due to the special size of A320/321, with a fuselage length of more than 40 meters, making it easy to distinguish. For some objects in other categories, such as ARJ21 and Boeing aircraft, due to their multi-scale characteristics and varied directions, this presents a challenge for the network in learning invariant features within the same category. We enhance the detection capabilities of multi-scale and directional information-rich objects by fusing global information and performing compensation in deep and shallow layers respectively, as well as learning object direction-aware information.

	\subsubsection{Comparison Results on SSDD}
	To confirm the superiority of our approach on different datasets, we selected the widely used SSDD dataset and compared it against several state-of-the-art object detection methods, some of which were specially designed for SAR imagery. A variety of experiments ensured the rigor of our work.

	As indicated in the data presented in Table \ref{tab:4.7}, our method exhibits strong competitiveness even when transferred to ship tasks. It can be observed that due to the relatively small size of the SSDD dataset, the current methods have already achieved a high level of average precision. However, due to the continued presence of multi-scale objects and interference from background noise, this has prevented further performance improvements in other methods. Note that our method excels in information fusion and transmission, making it less prone to influence by these problems and less prone to overfit on small datasets. Consequently, this further elevates the benchmark performance of the dataset. Additionally, in comparison to the state-of-the-art (SOTA) model YOLOv8's larger version, we still achieve a 0.4\% precision advantage on mAP$_{50}$, while reducing parameters and GFLOPS to one-fourth and one-sixth, respectively.

		\begin{figure*}[htpb]
		\centering
		\includegraphics[width=0.8\linewidth]{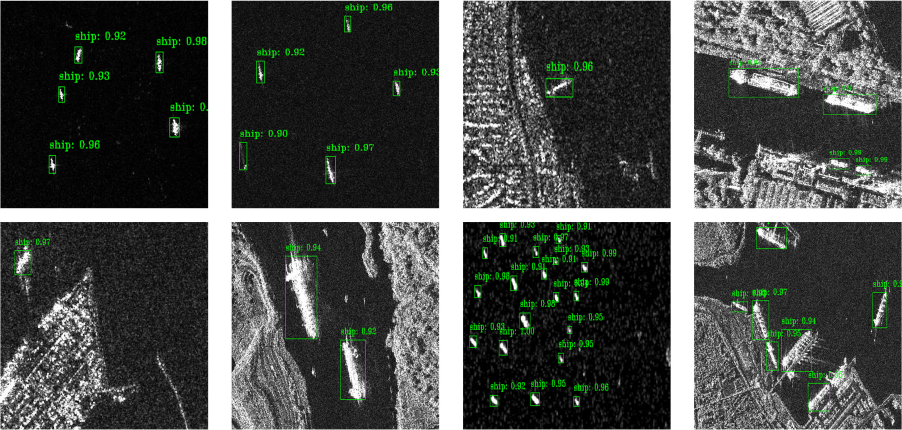}
		\caption{Visualization results of object detection under SAR-Net on SSDD.}
		\label{fig:4.3}
		\vspace{-0.1in}
	\end{figure*}
	\begin{figure*}[htpb]
		\centering
		\includegraphics[width=0.8\linewidth]{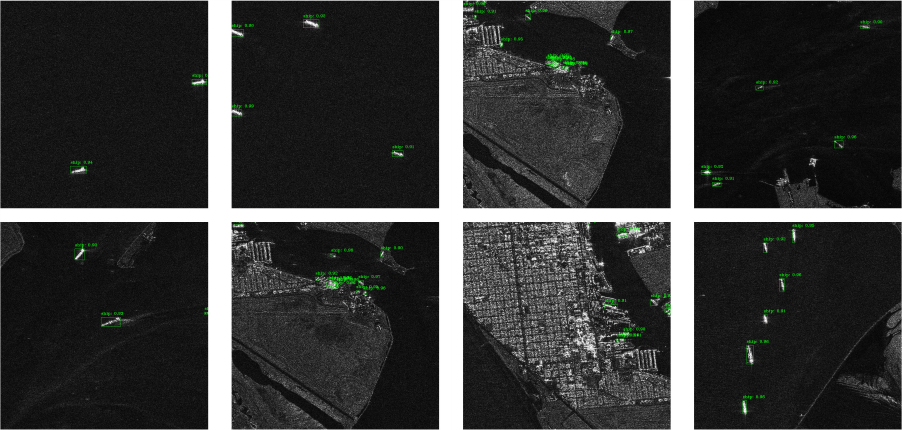}
		\caption{Visualization results of object detection under SAR-Net on HRSID.}
		\label{fig:4.4}
		\vspace{-0.1in}
	\end{figure*}
	\subsubsection{Comparison Results on HRSID}
	For additional validation of the generalization and robustness of our approach, we opted for the more challenging ship dataset, HRSID, in comparison to SSDD. HRSID presents greater difficulty with a higher number of instances, larger image resolutions, and significant variations in object scales. This dataset serves as a more stringent test of the model's capabilities in feature extraction and utilization.
	
	Continuing our evaluation, we chose state-of-the-art object detection methods, some of which were specially designed for SAR imagery. Compared with other approaches, our method maintains excellent performance with a relatively small computational cost. Consistent with the results from SSDD, when compared to YOLOv8m, our method exhibits a 0.6\% precision advantage on mAP$_{50}$ and significantly smaller parameter size and GFLOPs, being one-fourth and one-sixth of YOLOv8m, respectively.
	\subsubsection{Comparison Results on SSDD and HRSID with Oriented Bounding Box Detection Methods}
	To further explore the direction-aware attention module for detecting rotated objects, we compared it with state-of-the-art SAR image rotation detection methods and conducted experiments on two ship datasets with rotation annotations. The results, as illustrated in Table \ref{tab:4.9}, demonstrate that our method achieved the highest accuracy on both datasets. Despite not being specifically designed for rotation detection, the direction-aware attention module is still able to learn object feature information from two directions, thus assigning greater attention weights to the detected objects. Additionally, it retains this directional feature to provide perceptual information beneficial for detection.
	
	\subsection{Result Visualization}
	Due to the SAR-AIRcraft-1.0 dataset's nature of fine-grained object detection, it presents certain detection challenges, leading to notable variations in detection results among different methods. To provide a more intuitive representation of the performance of various methods on SAR-AIRcraft-1.0, we selected a few images and visualized the detection results using different methods, as illustrated in Fig. \ref{fig:4.2}. In comparison with other methods, our approach consistently identifies objects. While occasional false positives may occur, the incorporation of sufficient directional awareness information and the utilization of global fusion to minimize information loss result in minimal instances of false negatives in our detection results. For instance, the green bounding box (representing A220) in (b), the brown bounding box (representing ARJ21) in (c), and the orange bounding box (representing Boeing787) in (d) were undetected by other methods, highlighting the superiority of our approach in these scenarios.
	
	For the SSDD and HRSID datasets, given their high baseline levels of average precision and the single-category nature of object detection, we provide visualization results exclusively for SAR-Net. The visualizations for SSDD and HRSID are depicted in Fig. \ref{fig:4.3} and Fig. \ref{fig:4.4}. SAR-Net demonstrates excellent detection performance in different backgrounds.
	\section{Conclusion}
	As the focus on deep learning-based remote sensing tasks grows, SAR image object detection has become a crucial research direction. Establishing a complementary relationship between high-level semantic information and low-level detailed information without losing critical details is essential for solving the multi-scale object feature fusion problem. In response to this, we propose a framework capable of globally fusing multi-scale direction-aware information, named SAR-Net. This framework effectively addresses the challenges of information loss due to sequential local fusion and the difficulty in capturing detailed information under complex background interference, allowing multi-scale detection features to focus on global information. Experiments executed on SAR datasets of aircraft and ships demonstrate the effectiveness and excellent generalization of our framework. We believe this global information fusion perspective will provide valuable insights for SAR image detection tasks.
	
	\ifCLASSOPTIONcaptionsoff
	\newpage
	\fi
	{
		\bibliographystyle{IEEEtran}
		\bibliography{refs}
	}
	
\end{document}